\pdfoutput=1

\documentclass[11pt]{article}

\usepackage{emnlp2021}

\usepackage{times}
\usepackage{latexsym}
\usepackage{booktabs}
\usepackage[T1]{fontenc}

\usepackage[utf8]{inputenc}

\usepackage{microtype}
\usepackage{amsmath}
\usepackage{booktabs}
\usepackage{array}
\newcolumntype{L}{>{$}l<{$}}
\newcolumntype{C}{>{$}c<{$}}
\newcolumntype{R}{>{$}r<{$}}

\usepackage{times}
\usepackage{latexsym}

\usepackage{microtype}

\usepackage{times}
\usepackage{latexsym}

\usepackage[T1]{fontenc}

\usepackage[utf8]{inputenc}

\usepackage{multirow}
\usepackage{array, tabularx, caption, boldline}
\usepackage{graphicx}
\usepackage{cellspace}
\usepackage{algpseudocode}  
\usepackage{amsmath}
\usepackage{multicol}  
\usepackage{multirow} 
\usepackage{flexisym}
\usepackage{graphicx}  
\usepackage{array}

\usepackage{microtype}

\usepackage{tabularx}
\makeatletter
\def\hlinewd#1{%
\noalign{\ifnum0=`}\fi\hrule \@height #1 %
\futurelet\reserved@a\@xhline}
\makeatother

\usepackage{times}
\usepackage{latexsym}

\usepackage{algpseudocode}  
\usepackage{amsmath}
\usepackage{multicol}  
\usepackage{multirow} 
\usepackage{graphicx}  
\usepackage{array}
\usepackage{arydshln}
\usepackage{booktabs}
\usepackage{textcomp}

\usepackage{amssymb}
\usepackage{pifont}

\usepackage{cleveref}
\crefformat{section}{\S#2#1#3} 
\crefformat{subsection}{\S#2#1#3}
\crefformat{subsubsection}{\S#2#1#3}

\usepackage{microtype}

\usepackage{adjustbox}
\usepackage{multicol}  
\usepackage{multirow} 
\usepackage{amsmath}
\usepackage{amsfonts}
\usepackage{makecell}
\usepackage{algpseudocode} 
\usepackage{verbatim}
\usepackage{mathtools}
\DeclareMathOperator*{\argmax}{arg\,max}

\usepackage{booktabs}

\usepackage{makecell}
\usepackage{cleveref}
\usepackage[normalem]{ulem}

\usepackage{times}
\usepackage{latexsym}

\usepackage{footnote}
\makesavenoteenv{tabular}
\makesavenoteenv{table}

\usepackage[hang,flushmargin]{footmisc}

\usepackage[linesnumbered,ruled]{algorithm2e}
\SetAlFnt{\small}
\SetAlCapNameFnt{\normalsize}
\makeatletter
\newcommand{\nosemic}{\renewcommand{\@endalgocfline}{\relax}}
\newcommand{\dosemic}{\renewcommand{\@endalgocfline}{\algocf@endline}}
\let\oldnl\nl
\newcommand{\nonl}{\renewcommand{\nl}{\let\nl\oldnl}}
\makeatother

\usepackage{enumitem}
\usepackage{tablefootnote}

\usepackage{tabularx}
\makeatletter
\def\hlinewd#1{%
\noalign{\ifnum0=`}\fi\hrule \@height #1 %
\futurelet\reserved@a\@xhline}
\makeatother

\title{Few-Shot Table-to-Text Generation with Prototype Memory}

\author{Yixuan Su\quad  Zaiqiao Meng\quad Simon Baker\quad  
 Nigel Collier\quad\\

Language Technology Lab, University of Cambridge \\

{\tt \{ys484,zm324,sb895,nhc30\}@cam.ac.uk}\\
}

\begin{document}
\maketitle
\begin{abstract}
Neural table-to-text generation models have achieved remarkable progress on an array of tasks. However, due to the data-hungry nature of neural models, their performances strongly rely on large-scale training examples, limiting their applicability in real-world applications. To address this, we propose a new framework: Prototype-to-Generate (P2G), for table-to-text generation under the few-shot scenario. The proposed framework utilizes the retrieved prototypes, which are jointly selected by an IR system and a novel prototype selector to help the model bridging the structural gap between tables and texts. Experimental results on three benchmark datasets with three state-of-the-art models demonstrate that the proposed framework significantly improves the model performance across various evaluation metrics.
\end{abstract}

\section{Introduction}
Generating natural language from structured table \cite{DBLP:journals/jair/GattK18}, i.e. table-to-text generation, is an important research problem for various NLP applications, such as biographical descriptions \cite{DBLP:conf/emnlp/LebretGA16}, restaurant information \cite{DBLP:conf/sigdial/NovikovaDR17}, basketball game summaries \cite{DBLP:conf/emnlp/WisemanSR17}, and open-domain question answering \cite{DBLP:journals/corr/abs-2010-10439}. 

The main challenge of table-to-text generation stems from the structural difference between the table and the natural language text. With recent advances in neural networks, many sophisticated neural models \cite{DBLP:conf/aaai/LiuWSCS18,DBLP:conf/inlg/GehrmannDER18,DBLP:conf/aaai/Puduppully0L19,DBLP:conf/acl/PuduppullyDL19,su2021plangen} have been proposed to address this problem. While achieving impressive results, such neural models are data-hungry, i.e. large amounts of training data are required for them to learn the mapping between tables and texts. This can prohibit these models from being applied to real-world applications due to the huge data curation overhead \cite{DBLP:conf/acl/ChenECLW20}. This motivates us to investigate \textit{few-shot table-to-text generation} \cite{DBLP:conf/acl/MaYLLZS19,DBLP:conf/acl/ChenECLW20}, that allows the model to learn a satisfactory table-to-text mapping with limited labelled training data.
\\\indent In this work, we propose to address this problem by augmenting data-to-text generation models with prototype memory acquired from a large unlabelled corpus. Our motivation is two-fold: (1) Relevant human-authored texts, termed ``prototypes'', are informative and can teach the model how to better describe the table when limited training data is available. (2) However, traditional lexical-based IR systems, e.g. BM25, are inaccurate and the quality of their results are not guaranteed. Therefore, a BERT-based prototype selector is required to further select the prototypes, from the results retrieved by the IR system, that are closely related to the table for better guiding the neural generation model.
\begin{figure*}[t] 
	\centering    
	\setlength{\abovecaptionskip}{3pt}
\includegraphics[width=0.98\textwidth]{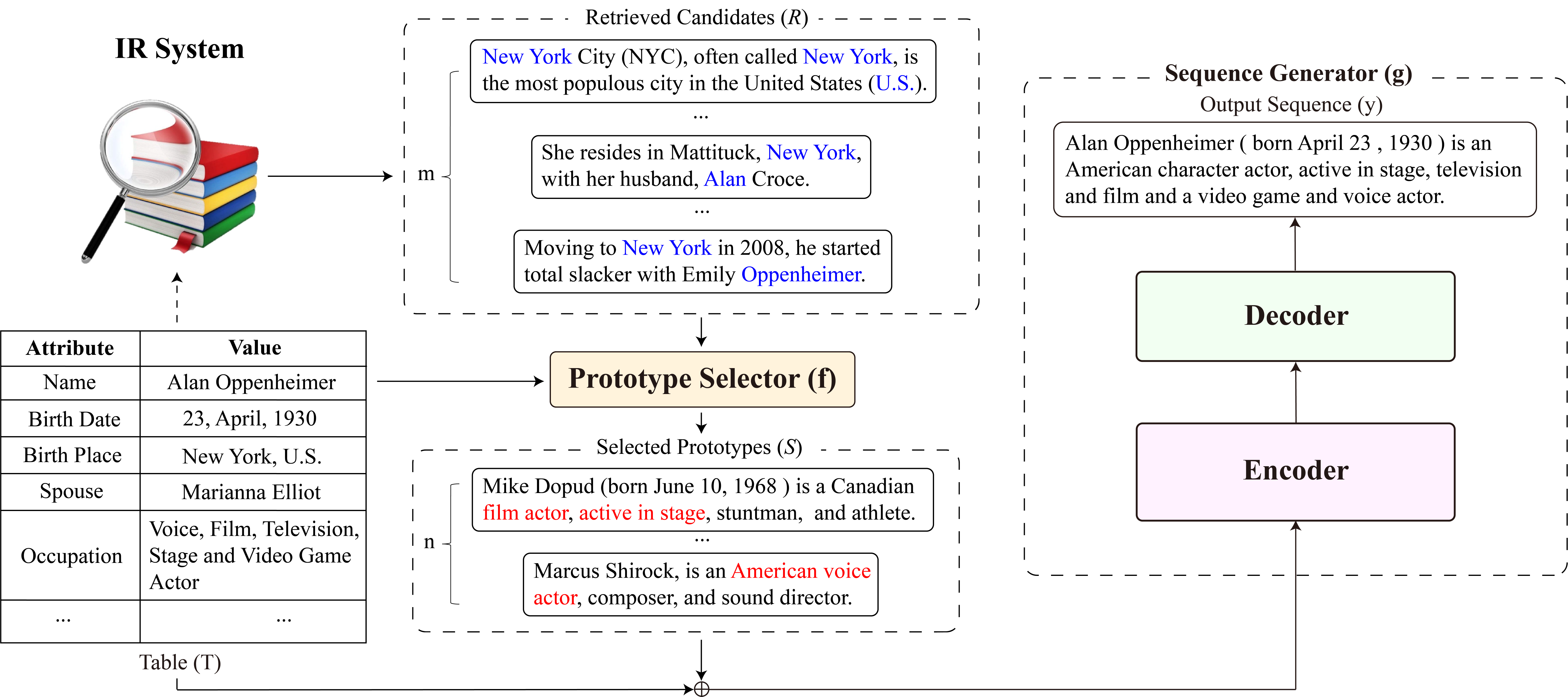}
    \caption{An overview of the proposed Prototype-to-Generate (P2G) framework.} 
    \label{fig:overview}
\end{figure*}
\\\indent Figure~\ref{fig:overview} illustrates the proposed Prototype-to-Generate (P2G) framework. Given the table, an IR system is first applied to retrieve candidates that are potentially related to the table from a large unlabelled corpus. Based on the retrieved candidates, a prototype selector then selects the top $n$ prototypes based on the table-text pairwise similarity.  Lastly, a sequence generator takes the table and the selected prototypes as input to produce the output. To prevent the model from uncritically copying the information contained in the prototypes that is irrelevant to the table, we introduce a content-aware learning objective when training the generator. \\\indent In recent years, retrieval-based (i.e. template-based) text generation has been studied in different NLP areas, including machine translation \cite{DBLP:journals/corr/GuWCL17}, unconditional text generation \cite{DBLP:journals/tacl/GuuHOL18}, dialogue systems \cite{DBLP:conf/aaai/0006WHWL019,DBLP:journals/taslp/SuWCBKC21}, paraphrase generation \cite{DBLP:conf/acl/KazemnejadSB20,DBLP:conf/acl/SuVBWC21}, and question answering \cite{DBLP:conf/nips/LewisPPPKGKLYR020}. 
Despite their differences, we identify two major limitations in previous studies compared to our approach. Firstly, most previous research \cite{DBLP:journals/corr/GuWCL17,DBLP:conf/aaai/0006WHWL019,DBLP:conf/acl/KazemnejadSB20} build their retrieval corpus based on data consisting of aligned source-target pairs, which precludes the use of abundant unlabelled data. Secondly, current retrieval mechanisms are either based on lexical similarity (e.g. BM25) where its accuracy cannot be guaranteed, or large neural networks \cite{DBLP:conf/emnlp/KarpukhinOMLWEC20} which require a large amount of data to train. \\\indent Notably, our framework is independent of the choice of generation model. For a comprehensive evaluation, we test our approach on three representative models, including the current state of the art. The experimental results on three datasets show that our framework leads to remarkable performance improvements across all evaluation metrics.

\begin{table*}[t]
    \small
	\centering  
	\renewcommand{\arraystretch}{1.2}
	\setlength{\tabcolsep}{3.2pt}
	\scalebox{0.81}{
	\begin{tabular}{ccccccccccccc}
		\hlinewd{0.75pt}
		Domain&\multicolumn{4}{c}{\textbf{Humans}}&\multicolumn{4}{c}{\textbf{Books}}&\multicolumn{4}{c}{\textbf{Songs}}\\
		\cmidrule(lr){1-1}
		\cmidrule(lr){2-5}
        \cmidrule(lr){6-9}
        \cmidrule(lr){10-13}
		Training Size&50&100&200&500&50&100&200&500&50&100&200&500\\
		\hlinewd{0.75pt}
		\textit{Retrieval-Based}&&&&&&&&&&&&\\
		\hline
		Retri-Gen&7.4/0.7&10.3/1.6&13.2/2.7&16.5/4.1&12.1/1.8&13.2/2.0&14.7/2.4&15.9/3.3&13.4/2.7&14.3/3.1&16.2/4.3&17.7/4.9\\
		RA-Gen&29.4/15.8&33.6/18.9&40.1/26.7&44.3/30.9&34.7/22.2&35.7/22.9&37.4/24.9&40.9/28.3&34.9/24.8&36.4/26.1&39.0/29.2&42.1/31.7\\
		\hline
		Struct-Aware$^\ddagger$&2.9/0.1&5.1/0.4&6.1/0.8&8.3/1.5&7.3/1.7&6.8/
		1.5&7.8/2.1&8.8/2.4&10.4/4.1&12.0/5.1&11.6/4.7&13.1/5.8\\
		Pivot&14.9/3.2&18.7/6.9&25.3/14.1&29.8/17.3&23.1/10.7&24.9/13.3&27.0/15.2&29.8/18.1&26.2/14.7&28.0/16.2&29.2/17.7&31.7/20.0\\
		KGPT&30.2/18.8&35.0/22.8&38.9/26.1&43.7/30.4&35.3/24.2&37.4/25.8&38.4/26.7&42.0/29.2&37.9/28.3&39.8/30.1&40.3/30.5&42.9/33.0\\
		Switch-GPT$^\dagger$&25.7/14.1&29.5/16.2&36.1/22.1&41.7/28.3&34.3/22.5&36.2/23.1&37.9/25.0&40.3/27.6&36.1/26.2&37.2/28.6&39.4/30.1&42.2/32.6\\
		Table-GPT$^\ddagger$&29.8/16.3&34.5/20.6&40.6/27.6&45.6/32.4&35.1/24.0&37.3/25.4&38.5/26.7&41.6/28.9&36.7/27.1&37.8/29.4&39.3/30.6&42.3/32.8\\
		T5-Prefix&32.6/20.7&37.1/23.1&41.7/28.8&46.3/33.2&34.2/21.2&38.3/26.7&39.4/27.6&42.9/30.0&37.6/28.1&38.7/29.2&40.0/30.3&43.5/33.9\\
		\hline
		P2G+Switch-GPT&31.4/19.9&36.5/22.7&42.0/30.1&45.8/32.6&38.2/25.4&39.9/27.3&41.7/29.2&44.6/31.7&39.1/29.9&40.3/30.7&41.8/32.0&45.0/35.4\\
		P2G+Table-GPT&34.9/23.2&38.9/25.1&43.1/31.2&48.1/35.0&40.1/29.3&41.0/28.6&43.1/30.4&47.0/34.0&41.2/31.7&42.7/33.6&44.2/34.9&47.9/38.1\\
		P2G+T5-Prefix&\textbf{39.3}/\textbf{27.9}&\textbf{42.6}/\textbf{30.8}&\textbf{46.2}/\textbf{34.0}&\textbf{50.1}/\textbf{37.3}&\textbf{41.2}/\textbf{28.3}&\textbf{43.4}/\textbf{30.5}&\textbf{46.4}/\textbf{33.8}&\textbf{49.2}/\textbf{36.1}&\textbf{42.8}/\textbf{33.0}&\textbf{45.9}/\textbf{35.7}&\textbf{47.6}/\textbf{37.5}&\textbf{50.7}/\textbf{40.1}\\
		\hlinewd{0.75pt}
	\end{tabular}}
    \caption{Results on datasets from three domains. In each entry, $x$/$y$ denotes the model performance on BLEU-4/ROUGE-4(F-measure). $^\dagger$ and $^\ddagger$ results are copied from \citet{DBLP:conf/acl/ChenECLW20} and \citet{DBLP:conf/coling/GongSFQBLL20}. All results acquired with the proposed framework outperform the original model with a significance level $p\textup{-value} < 0.01$.}
	\label{tb:result}
\end{table*}

\section{Methodology}
Figure~\ref{fig:overview} depicts an overview of our framework. Given a  linearized table $T=\{t_1, ..., t_{|T|}\}$, where $t_{i}=\{a_i, v_i\}$ is an attribute-value pair, an IR system first retrieves a set of $m$ candidates $\mathcal{R}$ from the large unlabelled corpus. Then, a prototype selector $f$ (\S \ref{sec:selector}) selects the top $n$ prototypes $\mathcal{S}$ from $\mathcal{R}$ that are most related to $T$. Lastly, a sequence generator $g$ (\S \ref{sec:generator}) takes $T$ and $\mathcal{S}$ to produce the output $y$.

\subsection{Prototype Selector}
\label{sec:selector}
As illustrated in Figure~\ref{fig:overview}, given the table $T$, the IR system relies on lexical features (e.g., word overlaps between the table and texts as colored in \textcolor{blue}{blue}) to retrieve candidates $\mathcal{R}$. However, such lexical features are inaccurate and the semantic relevance between $T$ and $\mathcal{R}$ cannot be guaranteed. To remedy this problem, we utilize a prototype selector $f$ to select the top $n$ prototypes $\mathcal{S}$ from $\mathcal{R}$ based on the table-text pairwise similarity. Formally, given the table $T$ and a text $r\in \mathcal{R}$, their pairwise similarity score is defined as $f(T, r)$ and $\mathcal{S}$ is then defined as:
\begin{equation}
\label{eq:s_definition}
    \mathcal{S}=\argmax_{\mathcal{R}^{\prime}\in \mathcal{R}, |\mathcal{R}^{\prime}|=n}\sum_{r\in \mathcal{R}^{\prime}}f(T, r).
\end{equation}
Figure~\ref{fig:overview} shows examples of the selected prototypes, $\mathcal{S}$.  
We see that $\mathcal{S}$ are 
better related to the table and being closer to the reference text, i.e., the reference and $\mathcal{S}$ could share similar contexts like the words in \textcolor{red}{red}. Thus, $\mathcal{S}$ can be deemed as an guiding signal which teaches the model how to describe the table.

In this work, we use BERT \cite{DBLP:conf/naacl/DevlinCLT19} to build the prototype selector. The score $f(T, r)$ is computed by a linear projection over the average embeddings of $\textup{BERT}([T\textup{:}r])$, where $[:]$ denotes concatenation operation. During training, given the table $T$, the reference text $y$, and the retrieved candidate set $\mathcal{R}$ provided by the IR system, the learning objective of the prototype selector is defined as:
\setlength{\belowdisplayskip}{0pt} \setlength{\belowdisplayshortskip}{0pt}
\setlength{\abovedisplayskip}{0pt} \setlength{\abovedisplayshortskip}{0pt}
\begin{equation}
\label{eq:neural_selector}
    \mathcal{L}_f = \sum_{j=1}^k\max\{0,1 - f(T, y) + f(T, \mathcal{R}_{j})\},
\end{equation}
where $\mathcal{R}_{j}\in\mathcal{R}$ and $k$ is the number of negatives sampled from $\mathcal{R}$. After training $f$, we can obtain the prototype-augmented dataset $\mathcal{D}=\{(T, \mathcal{S}, y)_i\}_{i=1}^{|\mathcal{D}|}$ for the learning of the generator.

\subsection{Sequence Generator}
\label{sec:generator}
The proposed framework is model-agnostic, thus the generator $g$ can be any generation model. Given a training example $(T,\mathcal{S},y)\in\mathcal{D}$, the learning of $g$ is defined as: 
$\mathcal{L}_{\textup{LM}} = -\sum_{i=1}^{|y|}\log p_{\theta}(y_i|y_{<i}; X)$,
where $\theta$ denotes the parameters of the generator, and $X=[T\textup{:}\mathcal{S}]$. Moreover, we introduce a new content-aware learning objective. Our motivation is that the prototypes $\mathcal{S}$ is likely to contain information that is irrelevant to the table, thus the generator should learn to ignore the irrelevant part of $\mathcal{S}$ and only focus on the useful information. To this end, inspired by \citet{DBLP:conf/iclr/WelleckKRDCW20}, we formulate the content-aware learning objective as: $\mathcal{L}_{\textup{CA}} = -\sum_{i=1}^{|y|}\sum_{\Tilde{y}\in \mathcal{S},  \Tilde{y}\notin y}\log(1-p_{\theta}(\Tilde{y}|y_{<i};X))$
which discourages the generation of the irrelevant tokens contained in $\mathcal{S}$. The generator overall learning objective is then defined as: $\mathcal{L}_{g}=\mathcal{L}_{\textup{LM}} + \mathcal{L}_{\textup{CA}}$.

\section{Experiment}
\subsection{Experiment Setup}
We conduct experiments on three benchmark few-shot table-to-text datasets \cite{DBLP:conf/acl/ChenECLW20} from different domains: \textit{Humans}, \textit{Books}, and \textit{Songs}. Following previous studies \cite{DBLP:conf/acl/ChenECLW20,DBLP:conf/coling/GongSFQBLL20}, we train our model on different settings by varying the training size from $\{50, 100, 200,500\}$, and evaluate our model using BLEU \cite{DBLP:conf/acl/PapineniRWZ02} and ROUGE \cite{lin-2004-rouge} metrics. Test sets of \textit{Humans}, \textit{Books}, and \textit{Songs} contain 13587, 5252 and 11879 instances.  \\\indent To build the IR system, we use Lucene\footnote{https://lucene.apache.org/core/} to pre-index all sentences contained in the English Wikipedia (Dec. 2018 dump). For each table, the IR system retrieves 100 sentences as the candidates $\mathcal{R}$. The prototype selector then select the top 3 results from $\mathcal{R}$ as the prototypes $\mathcal{S}$\footnote{To avoid the data leakage problem, when building the dataset, we make sure the prototypes do not contain the reference.}. When training the prototype selector, we set $k$ in Eq. \eqref{eq:neural_selector} as 5.\\\indent We compare our approach with both existing table-to-text methods that are not retrieval-based and also with the existing retrieval-based methods which we adapt for our concerned task. The existing table-to-text methods include Struct-Aware \cite{DBLP:conf/aaai/LiuWSCS18}, Pivot \cite{DBLP:conf/acl/MaYLLZS19}, Switch-GPT \cite{DBLP:conf/acl/ChenECLW20}, KGPT \cite{DBLP:conf/emnlp/ChenSYW20}, Table-GPT \cite{DBLP:conf/coling/GongSFQBLL20}, and T5-Prefix \cite{DBLP:journals/corr/abs-2007-08426}. The latter four are based on pre-trained language models (PLMs). 
The retrieval-based approaches include Retri-Gen \cite{DBLP:conf/aaai/0006WHWL019} and RA-Gen \cite{DBLP:conf/nips/LewisPPPKGKLYR020}, where RA-Gen is based on PLMs. We select three representative models (Switch-GPT, Table-GPT, and T5-Prefix) to test the proposed framework. 

\subsection{Main Results}
Table~\ref{tb:result} lists the experiment results, where P2G+X indicates using  model X under our framework. We can see that the proposed framework consistently and significantly improves the performance of all three models on all metrics, showing the robustness and universality of our approach. The notable performance gains suggest that the incorporation of retrieved prototypes greatly benefit the model's ability in bridging the gap between tables and texts. It is worth noting that the RA-Gen model applies a strong BART  \cite{DBLP:conf/acl/LewisLGGMLSZ20} as the generator. However, their retrieval module is purely based on a large neural models \cite{DBLP:conf/emnlp/KarpukhinOMLWEC20} that requires a large amount of data to train, and its accuracy degenerates when training data is limited, leading to the reduced generation performance.

\begin{table}[t]
    \small
	\centering  
	\renewcommand{\arraystretch}{1.2}
	\setlength{\tabcolsep}{6pt}
	\scalebox{0.85}{
	\begin{tabular}{ccccc}
		\hlinewd{0.75pt}
		Training Size&50&100&200&500\\
		\hlinewd{0.75pt}
		T5-Prefix&32.6/20.7&37.1/23.1&41.7/28.8&46.3/33.2\\
	    +\textit{Ret}&32.9/21.2&37.4/23.5&42.1/29.0&46.7/33.4\\
	    +\textit{Ret\&PS}&38.8/27.0&42.0/30.2&45.8/33.5&49.2/36.6\\
		+\textit{Ret\&PS\&CA}&\textbf{39.3}/\textbf{27.9}&\textbf{42.6}/\textbf{30.8}&\textbf{46.2}/\textbf{34.0}&\textbf{50.1}/\textbf{37.3}\\
		\hlinewd{0.75pt}
	\end{tabular}}
    \caption{Ablation study results on \textit{Humans} dataset. In each entry, $x$/$y$ denotes the BLEU-4/ROUGE-4 scores.}
	\label{tb:ablation_study}
\end{table}

\begin{figure}[t] 
	\centering    
	\setlength{\abovecaptionskip}{3pt}
\includegraphics[width=0.4\textwidth]{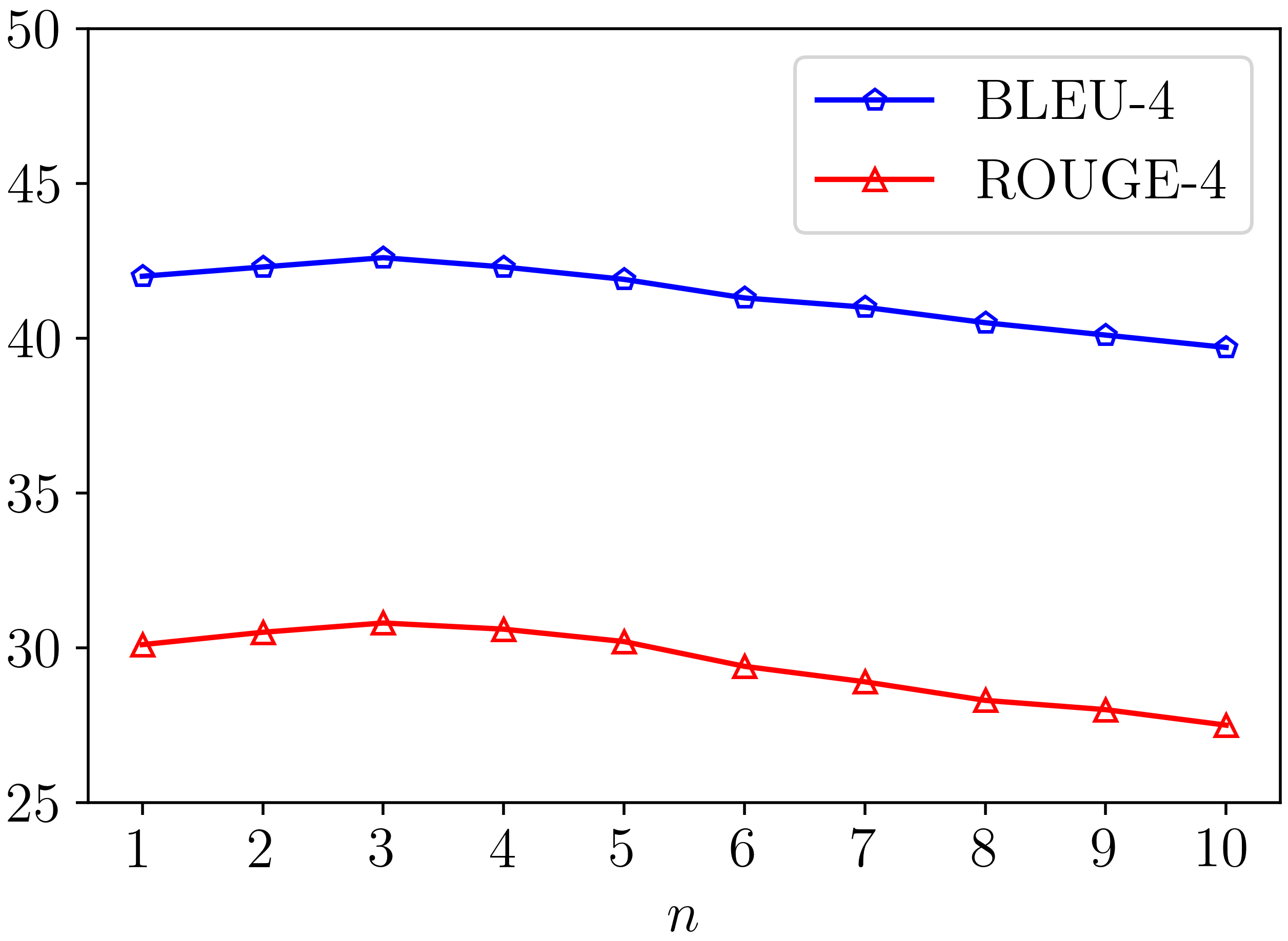}
    \caption{Effect of the number of prototypes ($n$).}
    \label{fig:n_effect}
\end{figure}

\subsection{Further Analysis}
In this section, we present further discussions and empirical analysis of the proposed model.
\paragraph{Ablation Study.} First, we perform ablation analysis on the T5-Prefix model by progressively incorporating each proposed technique. The +\textit{Ret} model directly utilizes the top $3$ retrieved results from the IR system as input. The +\textit{Ret\&PS} model utilizes the prototypes selected by the prototype selector as input. Finally, we include the proposed content-aware objective (+\textit{Ret\&PS\&CA}) which results in the same model as P2G+T5-Prefix. The experiments are conducted on the \textit{Humans} dataset with different training size. Table~\ref{tb:ablation_study} lists the results which show that each component positively contributes to the overall performance. By comparing T5-Prefix with +\textit{Ret}, we only observe a marginal improvement,  
suggesting that the retrieved results from the IR system are inaccurate  (i.e., unrelated to the table) which brings little help to the generator. Next, from the results of +\textit{Ret\&PS} model we see that the incorporation of prototype selector significantly boosts the performance. This is inline with our hypothesis that the prototype selector can select more accurate (i.e., related to the table and similar to the reference) prototypes that can effectively teach the generator about how to describe the table. Lastly, the results of +\textit{Ret\&PS\&CA} show that the proposed content-aware learning objective also benefits the model performance.

\paragraph{Effect of the Number of Prototypes.} Next, we examine how the number of prototypes ($n$ in Eq. \eqref{eq:s_definition}) affects the model performance. To this end, we train P2G+T5-Prefix with 100 instances on the \textit{Humans} dataset by varying the size of $n$. Figure~\ref{fig:n_effect} depicts the results of BLEU and ROUGE. We observe that, when $n$ is small (i.e., $n\leq 3$), the model performances are relatively the same. However, as $n$ approaching $10$, the results drop notably. The reason is that, as $n$ increases, the top $n$ prototypes are likely to contain more information that is irrelevant to the table (i.e. noisy information), which leads to the degeneration of model performances. 

\begin{table}[tb]
    \small
	\centering  
	\renewcommand{\arraystretch}{1.2}
	\scalebox{0.9}{
	\begin{tabular}{cccc}
		\hlinewd{0.75pt}
        &\textbf{\#Support}$\uparrow$&\textbf{\#Contradict}$\downarrow$&\textbf{Fluency}$\uparrow$\\
        \hline
        Agreement&0.64&0.61&0.53\\
        \hline
        Reference&4.27&0.31&1.85\\
        \hline
        Switch-GPT&3.23&0.98&1.37\\
        Table-GPT&3.47&0.75&1.42\\
        T5-Prefix&3.59&0.62&1.58\\
        \hline
        P2G+T5-Prefix&\textbf{3.98}&\textbf{0.47}&\textbf{1.71}\\
		\hlinewd{0.75pt}
	\end{tabular}}
    \caption{Human Evaluation Results. $\uparrow$ means the higher the better and $\downarrow$ means the lower the better. 
    }
	\label{tb:human_evaluation}
\end{table}

\begin{table*}[tb]
    \small
	\centering  
	\renewcommand{\arraystretch}{1.4}
	\scalebox{0.93}{
	\begin{tabular}{cl}
	\hlinewd{0.75pt}
    Table:&\makecell[l]{\textbf{Name}[The Absence]	\textbf{Background} [Grouporband]	\textbf{Origin}[Tampa, Florida, U.S.]	\\\textbf{Genre}[ melodic death metal, thrash metal] \textbf{Years Active}[2002--present]	\\	\textbf{Current Members}[Jamie Stewart, Patrick Pintavalle, Mike Leon, Jeramie Kling, Per Nilsson]	\\\textbf{Past Members}[Justin Reynolds, Nicholas Calaci, John Allen, Chris Pistillo, Peter Joseph]}\\
    \hline
    Reference:&\makecell[l]{The Absence is an American melodic death metal band from Tampa, Florida.}\\
    \hline
    T5+Prefix:&\makecell[l]{Jamie Stewart Patrick Pintavalle Mike Leon Jeramie Kling Per Nilsson, current members\\ is Justin Reynolds Nicholas Calaci John Allen Chris Pistillo Peter Joseph.}\\
    \hline
    \multicolumn{2}{c}{P2G+T5-Prefix}\\
    \hline
    Prototypes:&\makecell[l]{\textbf{1}: One Man Army and the Undead Quartet was a Swedish \textcolor{red}{band}, that played a fusion of \textcolor{red}{melodic}\\\; \; \textcolor{red}{death metal} and thrash metal.\\
    \textbf{2}: Epoch of Unlight is a \textcolor{red}{melodic death metal band} from Memphis, Tennessee.\\
    \textbf{3}: Inactive Messiah is a Greek \textcolor{red}{melodic death metal band}, founded in Athens.}\\
    \hline
    Output:&The Absence is an American \textcolor{red}{melodic death metal band} from Tampa, Florida, U.S.\\
    \hline
    \hline
    Table:& \makecell[l]{\textbf{Name}[Axel Toupane] \textbf{Position}[shooting guard/small forward]	\textbf{Height ft}[6]	\textbf{Height in}[7]	\textbf{Weight lb}[197]\\	\textbf{League}[NBA]	\textbf{Team}[Toronto Raptors] \textbf{Nationality}[French] \textbf{Draft Year}[2014]	\textbf{Birth Date}[23 July 1992]\\	\textbf{Birth Place}[Mulhouse, France]		\textbf{Career Start}[2011]	\textbf{Years}[2011--2015]}\\
    \hline
    Reference:&\makecell[l]{Axel Toupane (born July 23, 1992) is a French professional basketball player who currently plays for the\\ Toronto Raptors of the National Basketball Association (NBA).}\\
    \hline
    T5+Prefix:&\makecell[l]{Axel Toupane (born 23 July 1992) is a French professional basketball player.}\\
    \hline
    \multicolumn{2}{c}{P2G+T5-Prefix}\\
    \hline
    Prototypes:&\makecell[l]{\textbf{1}: Shannon Scott (born December 21, 1992) is an American \textcolor{red}{professional basketball player} who currently\\ \; \; plays for the \textcolor{red}{Toronto Raptors}.\\
    \textbf{2}: Bismack Biyombo Sumba (born August 28, 1992) is a Congolese \textcolor{red}{professional basketball player} who\\ \;\; \;currently plays for the \textcolor{red}{Toronto Raptors} of the \textcolor{red}{National Basketball Association}.\\
    \textbf{3}: Jama Mahlalela (born in Swaziland) is an assistant coach for the \textcolor{red}{Toronto Raptors} of the \textcolor{red}{NBA}.}\\
    \hline
    Output:&\makecell[l]{Axel Toupane (born July 23, 1992) is a French \textcolor{red}{professional basketball player} in the team of the \textcolor{red}{Toronto}\\ \textcolor{red}{Raptors} of the \textcolor{red}{National Basketball Association} (\textcolor{red}{NBA}).}\\
	\hlinewd{0.75pt}
	\end{tabular}}
    \caption{Examples of generated result from \textit{Humans} dataset. (best viewed in color)}
	\label{tb:human_case_study}
\end{table*}

\subsection{Human Evaluation}
We also conduct a human evaluation to assess the P2G+T5-Prefix model against several strong baselines, using graders proficient in English from an internal grading platform. Experiments are conducted on \textit{Humans} dataset using 100 training instances and we randomly select 300 test cases for evaluation. All generated results, plus the reference, are evaluated by three graders on two aspects: (1) \textit{factual correctness}; and (2) \textit{language fluency}. Firstly, the graders are asked to  count how many facts contained in the output are consistent with the table (\#Support), and are contradicted to the table (\#Contradict). Secondly, the graders are asked to assess the output in terms of language fluency on a 3-point Likert scale (0, 1, or 2).

Table~\ref{tb:human_evaluation} lists the evaluation results, with the first row showing strong inter-annotator agreements as measured by Fleiss$\textprime$ kappa coefficient \cite{fleiss1971mns}. The results show that our model (P2G+T5-Prefix) significantly outperforms other baseline models on all metrics (Sign Test with p-value < 0.05). The performance gains of P2G+T5-Prefix over T5-Prefix further suggest that the prototypes help the model to produce not only more syntactically fluent but also more factually correct outputs. 

\section{Case Study}
In Table \ref{tb:human_case_study}, we present two generated examples from our model. For comparison, we also show the results generated by the strongest baseline (T5-Prefix) along with the reference sentence. As for our model, we show the selected prototypes along with the generated output. Both our model and the baseline model are trained with 100 instances. 

As seen in the first case, the T5-Prefix fails to produce a correct output which describes the band. Instead, it just elaborates the name of the band members based on the table. In contrast, by relying on the prototypes that are related to the table, our model (P2G+T5-Prefix) produces an output that properly describes the band. Similarly, in the second case, the result of our model is more diverse and contains more facts that are supported by the table. These results further demonstrate that the prototypes can be deemed as effective guiding signals which teach the model how to describe the table. For better illustration, we highlight the parts, with \textcolor{red}{red} color, of prototypes on which the model relies when producing the output.

\section{Conclusion}
In this study, we introduced a new retrieval-based framework, Prototype-to-Generate (P2G), which augments table-to-text models with prototype memory from unlabelled data. Extensive experiments and analysis on three benchmark datasets show that our approach can significantly improve the performance of various strong generation models on all  evaluation metrics. Our code, models and other related resources can be found in  \url{https://github.com/yxuansu/Few-Shot-Table-to-Text-Generation}

\section*{Acknowledgments}
The authors wish to thank our anonymous reviewers for their insightful suggestions and comments.


\bibliography{anthology,custom}
\bibliographystyle{acl_natbib}

\clearpage

\appendix
\section{Related Work}
\paragraph{Table-to-Text Generation.} Table-to-text generation is a long-standing problem \cite{DBLP:journals/nle/ReiterD97} that aims at producing natural language descriptions of structured table. Traditional systems are primarily built on template-based algorithms \cite{oh-rudnicky-2000-stochastic,stent-etal-2004-trainable,DBLP:conf/acl/KondadadiHS13}. With recent advances in neural networks, researchers have built different neural models based on various strategies, e.g. latent-variables \cite{DBLP:conf/emnlp/WisemanSR18,DBLP:conf/iclr/YeS0W020}, structure awareness \cite{DBLP:conf/aaai/LiuWSCS18,DBLP:conf/inlg/ColinG19}, copy mechanism \cite{DBLP:conf/inlg/GehrmannDER18,DBLP:conf/aaai/Puduppully0L19,DBLP:conf/acl/PuduppullyDL19}, and pre-trained language models (PLMs)  \cite{DBLP:conf/emnlp/ChenSYW20,DBLP:journals/corr/abs-2005-10433,DBLP:journals/corr/abs-2007-08426}. More recently, to alleviate the data-hungry nature of neural models, \citet{DBLP:conf/acl/MaYLLZS19} applied a pipeline model which first selects key facts from the table before producing the output. \citet{DBLP:conf/emnlp/ChenSYW20} designed a knowledge-grounded strategy for language model pre-training. \citet{DBLP:conf/acl/ChenECLW20} and \citet{DBLP:conf/coling/GongSFQBLL20} adapted the pre-trained GPT-2 model with different architectural designs, e.g. switch policy \cite{DBLP:conf/acl/ChenECLW20} and content matching \cite{DBLP:conf/coling/GongSFQBLL20}, to address the few-shot table-to-text generation problem. 

\paragraph{Retrieval-Based Text Generation.} In the last few years, retrieval-based text generation has attracted much attention. \citet{DBLP:journals/corr/GuWCL17} utilized a search engineer to assist the neural machine translation model. \citet{DBLP:journals/tacl/GuuHOL18} addressed unconditional text generation with a neural editor model that edits the retrieved prototypes. \citet{DBLP:conf/aaai/0006WHWL019} and \citet{DBLP:journals/taslp/SuWCBKC21} incorporated retrieval frameworks into Seq2seq models to enrich the information contained in the dialogue responses. \citet{DBLP:conf/acl/KazemnejadSB20} applied a retrieval model to assist the generation of paraphrased sentence.  \citet{DBLP:conf/nips/LewisPPPKGKLYR020} incorporated external knowledge using a retrieval model for knowledge-intensive question answering. To the best of our knowledge, our work is the first one which explores how retrieval-based approach could benefit neural models for table-to-text generation task.

\section{Human Evaluation Guidelines}
In the human evaluation, the graders are asked to assess the results from two aspects. Following previous research \cite{DBLP:conf/acl/ChenECLW20,DBLP:conf/coling/GongSFQBLL20}, in the first study, the graders evaluate the \textit{factual correctness} of the generated results by counting how many facts contained in the output are consistent with the table (\#Support), and are contradicted to the table (\#Contradict). In the second study, the graders assess the \textit{language fluency} of the generated results following a 3-point Likert scale (0, 1, or 2). The definitions of different scores are provided as following:
\begin{itemize}
    \item $2$: The result is grammatically fluent and is easy to understand.
    \item $1$: The result contains small errors but the errors does not affect your understanding.
    \item $0$: The result does not make sense and it is unreadable.
\end{itemize}

\end{document}